\documentclass{article}

\usepackage{PRIMEarxiv}

\usepackage[utf8]{inputenc} 
\usepackage[T1]{fontenc}    
\usepackage{hyperref}       
\usepackage{url}            
\usepackage{booktabs}       
\usepackage{amsfonts}       
\usepackage{nicefrac}       
\usepackage{microtype}      
\usepackage{lipsum}
\usepackage{fancyhdr}       
\usepackage{graphicx}       
\usepackage{enumitem} 
\usepackage{float}
\usepackage{amsmath}
\usepackage{cleveref}
\graphicspath{{media/}}     
 
\title{CryoSAMU: Enhancing 3D Cryo-EM Density Maps of Protein Structures at Intermediate Resolution with Structure-Aware Multimodal U-Nets}

\author{
  Chenwei Zhang \textsuperscript{1} \ \ \ \ \ \ \ \ \ \
  Khanh Dao Duc \textsuperscript{2} \\
  \textsuperscript{1} Department of Computer Science, UBC \;\;\;\;\;
  \textsuperscript{2} Department of Mathematics, UBC\\
  \texttt{cwzhang@cs.ubc.ca} \;\;\;\;\;\;\;\;\;\;\;\;\;\;
  \texttt{kdd@math.ubc.ca}\\
}

\begin{document}
\maketitle
 
 
\begin{abstract}
Enhancing cryogenic electron microscopy (cryo-EM) 3D density maps at intermediate resolution (4-8 {\AA}) is crucial in protein structure determination. Recent advances in deep learning have led to the development of automated approaches for enhancing experimental cryo-EM density maps. Yet, these methods are not optimized for intermediate-resolution maps and rely on map density features alone. To address this, we propose CryoSAMU, a novel method designed to enhance 3D cryo-EM density maps of protein structures using structure-aware multimodal U-Nets and trained on curated intermediate-resolution density maps. We comprehensively evaluate CryoSAMU across various metrics and demonstrate its competitive performance compared to state-of-the-art methods. Notably, CryoSAMU achieves significantly faster processing speed, showing promise for future practical applications. Our code is available at \url{https://github.com/chenwei-zhang/CryoSAMU}.
\end{abstract}


\section{Introduction}
\label{sec:intro}

Cryogenic electron microscopy (Cryo-EM) has become one of the most prevalent techniques in structural biology for determining protein structures, thereby accelerating structure-based drug discovery~\cite{dd1,dd2}. Cryo-EM projects a series of 2D images, which are then reconstructed into 3D electron density maps, providing voxelized representations of proteins. While cryo-EM 3D maps serve as the basis for molecular structure determination, using raw maps is usually not possible as they often lack contrast due to various factors, including molecular motion and heterogeneity, imaging artifacts, and incoherent averaging of image data~\cite{eminherent}. To address these limitations, various approaches have been developed to enhance map quality by sharpening or modifying map densities~\cite{phenixautosharpen,relionpostprocess,LocalDeblur,deepemhancer,cryofem,emgan,emready}. 
Traditional methods rely on B-factor correction, which can be applied globally~\cite{phenixautosharpen,relionpostprocess} and locally~\cite{LocSpiral,LocalDeblur}. However, these methods struggle with maps exhibiting varying signal-to-noise ratios and lacking prior knowledge (e.g. local resolution) \cite{emready}.

With recent advancements in deep learning (DL), fully data-driven methods have been developed to automatically enhance raw cryo-EM maps for protein structure modeling. Leveraging neural networks such as convolutional neural networks (CNNs)~\cite{cnn}, generative adversarial networks (GANs)~\cite{gan}, and Transformers~\cite{transformer}, these methods achieved promising results in map enhancement. Yet, they are not optimized for intermediate-resolution maps (i.e., 4-8 {\AA}~\cite{emready}) and rely solely on a single modality---the density map itself---during neural network training, overlooking other relevant modalities such as structural information. This limitation restricts their ability to generalize across diverse protein structures and prevents them from fully leveraging complementary biological information. To address these shortcomings, we thus introduce \textbf{CryoSAMU}, a novel approach that combines 3D map features with structural embeddings derived from the pretrained protein language model ESM-IF1~\cite{esmif} to enhance 3D \textbf{Cryo}-EM density maps with \textbf{S}tructure-\textbf{A}ware \textbf{M}ultimodal \textbf{U}-Nets.

Our main contributions are:
\begin{itemize}
    \item We propose the first multimodal network that integrates structural information into a 3D U-Net model using cross-attention mechanisms for cryo-EM map enhancement.
    \item We develop a self-attention-based post-processing procedure for ESM-IF1's structural embeddings, effectively preserving both chain and residue relationships while maintaining structural integrity.
    \item We train CryoSAMU on a curated dataset of joint density maps at intermediate resolution and associated protein structures, optimizing it for map enhancement.
    \item We benchmark CryoSAMU against state-of-the-art approaches across various evaluation metrics over diverse tested samples. We achieve competitive level of performance but with significantly faster processing speeds (approximately 4.2 to 16.7 times), making our method well-suited for large-scale and practical applications.
    \item Our ablation study demonstrates significant improvement brought from integrating structural information.
\end{itemize}

\section{Related Work}
\label{relatedwork}

\subsection{Existing map enhancement methods}

Conventional map enhancement (sharpening) approaches, including Phenix Autosharpen~\cite{phenixautosharpen} and RELION postprocessing~\cite{relionpostprocess}, are based on global B-factor correction. This technique enhances the amplitude of high-frequency Fourier components in raw cryo-EM maps. However, global B-factor-based methods encounter difficulties with maps exhibiting heterogeneous local resolutions, often leading to over- or under-sharpening in specific regions. Despite local B-factor-based sharpening algorithms~\cite{LocSpiral,LocalDeblur} have been developed to alleviate this limitation, these methods still suffer from poor accuracy in estimating the local resolution of maps, which is crucial for precise local B-factor sharpening.

DeepEMhancer~\cite{deepemhancer} is a pioneering DL-based fully automatic method that leverages a 3D U-Net model to mimic local sharpening effects and enhance map features. Subsequently, CryoFEM~\cite{cryofem} that employs convolutional neural networks (CNNs) and EM-GAN~\cite{emgan} that utilizes generative adversarial networks (GANs) have been introduced to further enhance cryo-EM maps. Most recently, with the emergence of vision transformers, EMReady~\cite{emready}, which adopts a Swin transformer architecture~\cite{swintransformer}, has shown superior performance in enhancing map quality for accurate protein structure modeling.

\subsection{Protein large language models}
The advancement of protein large language models (pLLMs) has enabled unprecedented insights into protein structure, function, and evolution~\cite{esm1,esm2,esm3,esmif,proteinbert,prostt5,proteinmpnn}. In analogy to human texts, protein sequences are treated as ``biological texts'' and input into pLLMs to capture contextual information inherent in the sequences. Notable examples of such models include the ESM family~\cite{esm1,esm2,esm3}, which are pretrained on vast datasets of protein sequences using the masked language modeling strategy, allowing them to develop rich representations that encapsulate evolutionary information. 

Addressing the inverse problem of predicting protein sequences from given structures, ESM-IF1~\cite{esmif} has been developed. Trained on 12 million protein structures derived from AlphaFold2~\cite{alphafold2}, ESM-IF1 predicts protein sequences from backbone atom coordinates. It is specifically designed to encode both sequence and structural information, including backbone geometry, side chain orientations, and secondary structure elements. These traits make ESM-IF1 a compelling choice for generating structure-aware embeddings that complement the map-only modality.

\begin{figure*}[t]
  \centering
   \includegraphics[width=\linewidth, trim={0.6cm 14.6cm 3.2cm 0cm}, clip]{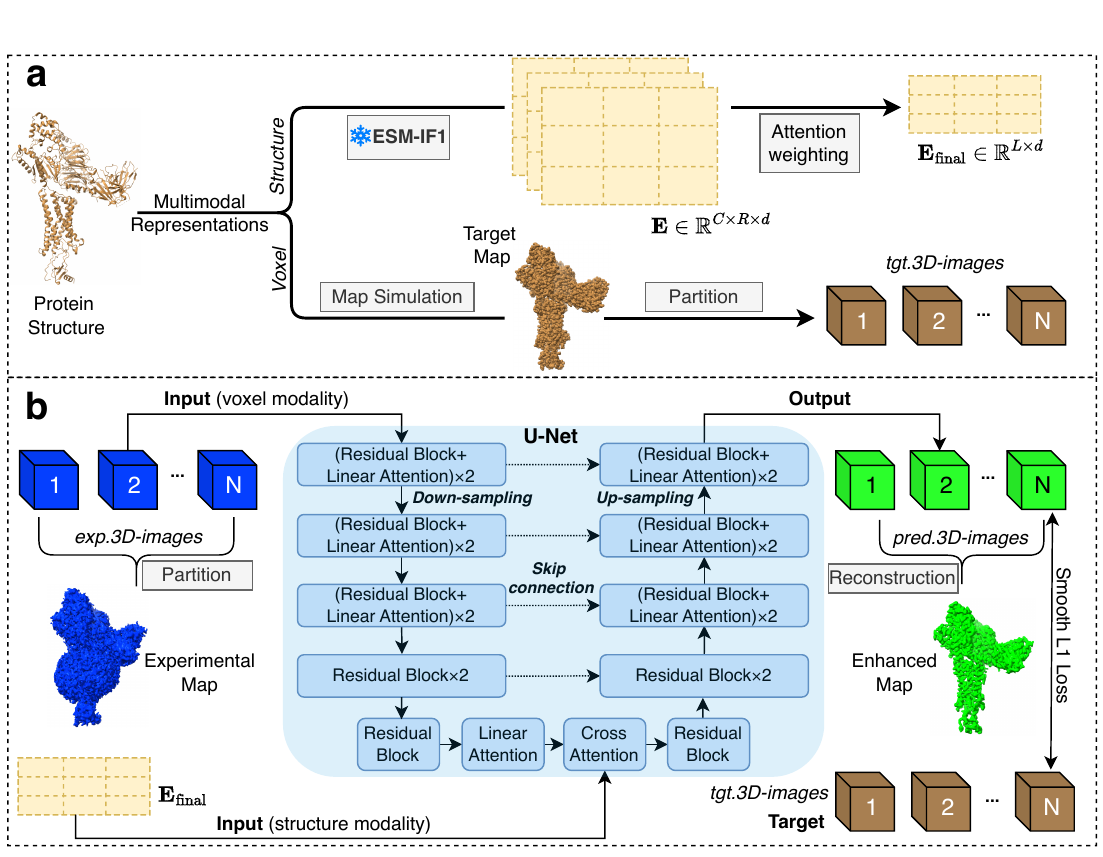}
    \vspace{-15pt}
   \caption{Overview of the CryoSAMU framework.
   \textbf{a} Generating protein multimodal representations: structure features are derived from a frozen pretrained ESM-IF1 model with self-attention weighting for a fixed-size representation; map voxel features are simulated via resolution-lowering point spread function and partitioned into smaller cubes. 
   \textbf{b} The CryoSAMU architecture. The experimental map is partitioned into smaller cubes and processed by a U-Net with residual blocks and linear attention modules. Structural embeddings are integrated into the bottleneck layer with cross-attention mechanism. The output cubes are reconstructed into the full-size enhanced map.
   }
   \label{fig:model}
\end{figure*}

\section{Method}

\subsection{Dataset of protein structures and density maps}

Our dataset was built with a set of cryo-EM density maps at resolutions from 4.0 {\AA} to 7.9 {\AA} from the EMDB databank~\cite{EMDB} and their associated protein structures from the PDB databank~\cite{PDB}. 
To ensure that density maps are properly aligned with their corresponding PDB structures, we excluded maps and PDBs from the dataset if:
(i) maps contain extensive regions without or misaligned corresponding PDB structures;
(ii) maps contain other macromolecules except proteins;    
(iii) PDB structures contain backbone atoms only and/or unknown residues.
Furthermore, to enhance training efficiency, we measured the correlation between map-PDB pairs using ChimeraX~\cite{chimerax}, and removed pairs with correlation score lower than 0.65. To avoid data redundancy, we measured the sequence identity between PDB structures, and retained only one if identity is greater than 30 \%.
As a result, a total of 384 pairs of cryo-EM maps and associated PDB structures remained. Among these data, 247 ($\sim$65 \%), 62 ($\sim$15 \%), and 75 ($\sim$20 \%) map-PDB pairs were selected as training, validation, and test sets, respectively. Details are listed in Supplementary Tables \ref{tab:train_data} to \ref{tab:test_data}.

\subsection{Multimodal representations of protein structures}
\noindent \textbf{Generating 3D target maps from protein structures} \hspace{3pt}
For input experimental maps (denoted as \emph{ExpMaps}) in training and validation sets, we simulated the corresponding target maps (denoted as \emph{TgtMaps}) from associated protein structures using the \texttt{StructureBlurrer} package in TEMPy2~\cite{tempy2}. The simulation was performed with a grid interval of 1 {\AA} and a resolution cutoff at 2 {\AA}, based on the convolution of atom points with resolution-lowering point spread functions. Given a PDB structure with $M$ atoms, the simulated density $\rho$ at grid point $x$ is calculated by:
\begin{equation}
    \rho(\mathbf{x}) = \sum_{i=1}^{M} \theta Z_i e^{-k|\mathbf{x}-\mathbf{r}_i|^2},
\end{equation}
where $Z_i$ and $\mathbf{r}_i$ refer to the atomic number and the position vector of the i-th heavy atom, respectively. Here, $\theta$ is a scaling factor and $k$ is defined based on resolution~\cite{emready,embuild}. 

\noindent \textbf{Resampling 3D maps} \hspace{3pt}
We first resampled both \emph{ExpMaps} and \emph{TgtMaps} to 1 {\AA}/voxel since the cryo-EM maps vary in voxel size. Subsequently, we normalized the density values to a range of 0 to 1 using the 99.9th percentile density value of each map. Due to GPU memory constraints, we partitioned \emph{ExpMaps} and \emph{TgtMaps} into smaller 3D subvolume pairs (denoted as \emph{exp.3D-images} and \emph{tgt.3D-images}) with size of $64\times64\times64$, the largest feasible size that allows for a sufficient batch size (See Figure \ref{fig:model}a.). To mitigate boundary artifacts during truncation, we applied zero-padding of 64 voxels on each side along all dimensions. As a result, a total of 29829 \emph{exp-tgt} image pairs were yielded for network training and 4642 for validation.

\noindent \textbf{Generating structural embeddings} \hspace{3pt}
We employed ESM-IF1~\cite{esmif} to generate protein structural embeddings, which will serve as an additional modality for network training. Specifically, we derived the embeddings by first extracting backbone coordinates (N, C$\alpha$, and C atoms) from a PDB file, ensuring that only standard residues with complete backbone information are included. We then fed these coordinates into ESM-IF1 to generate embeddings for each protein chain. Since the lengths of chains varied, we applied zero-padding to standardize the embeddings.

\noindent \textbf{Fixed-size representation with attention weighting} \hspace{3pt}
Following the generation of embeddings, we implemented self-attention weighting to create fixed-size representations while preserving the intrinsic relationships between chains and residues. To this end, we computed attention weights based on embedding similarity to identify the most informative regions. Specifically, given a PDB structure containing $C$ chains and $R$ residues per chain, its structural embedding derived from ESM-IF1 is denoted as $\mathbf{E} \in \mathbb{R}^{C\times R \times d}$, where $d=512$ is the embedding dimension. 
We carried out the refinement process in several steps. First, we computed chain-level embeddings by averaging across residues:
\begin{equation}
    \mathbf{E}_{\text{chain}} = \frac{1}{R} \sum_{j=1}^{R} \mathbf{E}_{:, j, :}, \quad \mathbf{E}_{\text{chain}} \in \mathbb{R}^{C \times d}.
\end{equation}
Next, we computed the similarity matrix to determine the relative importance of each chain:
\begin{equation}
    \mathbf{S} = \mathbf{E}_{\text{chain}} \cdot \mathbf{E}_{\text{chain}}^T, \quad \mathbf{S} \in \mathbb{R}^{C \times C},
\end{equation}
where each element $\mathbf{S}_{ij}$ represents the similarity between chain $i$ and chain $j$: $\mathbf{S}_{ij} = \mathbf{E}_{\text{chain},i} \cdot \mathbf{E}_{\text{chain},j}^T$.
To derive attention weights, we applied a column-wise softmax function to $\mathbf{S}$: 
\begin{equation}
    \mathbf{W}_{ij} = \frac{\exp(\mathbf{S}_{ij})}{\sum_{k=1}^{C} \exp(\mathbf{S}_{ik})}.
\end{equation}
We then aggregated these weights across chains to assign a single importance weight per chain:
\begin{equation}
    \mathbf{w}_i = \frac{1}{C} \sum_{j}^{C} \mathbf{W}_{ij}, \quad i = 1, 2, \dots, C.
\end{equation}
These weights $\mathbf{w} = [\mathbf{w}_1, \mathbf{w}_2, .., \mathbf{w}_C]$ reflect the relative importance of each chain, and we leveraged them to aggregate chain-level embeddings into a unified representation: 
\begin{equation}
    \mathbf{E}_{\text{pooled}} = \sum_{i=1}^{C} w_i \mathbf{E}_{i,:,:}, \quad \mathbf{E}_{\text{pooled}} \in \mathbb{R}^{R \times d}.
\end{equation}
We further measured the importance of each residue in $\mathbf{E}_{\text{pooled}}$ using a residue-level similarity matrix. Following the same procedure as the chain-level weighting, we obtained a scalar weight $\alpha_j$ for each residue $j$, where $j=1, 2, \dots, R$.
Finally, we applied min-max normalization and resampled the embedding $\mathbf{E}_{\text{pooled}}$ based on the attention weights to a fixed-size representation, $\mathbf{E}_{\text{final}} \in \mathbb{R}^{L \times d}$, where $L=800$. When the input length $R > L$, we selected the top-$L$ residues with the highest attention weights. Conversely, when $R < L$, we sorted the residues by their attention weights and repeated them $\lceil L/R \rceil$ times to reach the target length, ensuring each resulting embedding maintains rich representations and consistent dimensions.

\subsection{The model architecture}
We proposed a structure-aware multimodal 3D U-Net, as depicted in Figure \ref{fig:model}b. The network contains an encoder, bottleneck, and decoder, interconnected by skip connections.

\noindent \textbf{Encoder} \hspace{3pt}
The input to the encoder is a 3D volume with a single channel. The encoder comprises four hierarchical layers. The first three layers each consist of two residual blocks, with each block incorporating a group normalization, a SiLU activation~\cite{silu}, and a dropout (p=0.2), followed immediately by a linear self-attention module with 4 heads~\cite{linearattention} to capture long-range (global) dependencies across voxels. The channel depth progressively increases as features are abstracted. In the fourth layer, only residual blocks are employed, producing a higher-level feature representation without the addition of attention modules.

\noindent \textbf{Bottleneck} \hspace{3pt}
At the bottleneck layer, the feature representation is first refined by a residue block and then by a linear self-attention module. Subsequently, a cross-attention block is introduced to fuse and align the volumetric features with structural embeddings using multi-head attention with 4 heads, where queries are derived from the volume features and keys/values from structural embeddings. This process enables structural conditioning while preserving spatial relationships. A second residual block is then applied to further fuse the combined features from both modalities. 

\noindent \textbf{Decoder} \hspace{3pt}
The decoder follows a symmetric architecture to the encoder. Feature maps are progressively upsampled using nearest-neighbor interpolation combined with 3D convolutions, and skip connections incorporate corresponding features from the encoder. Finally, a group normalization, a SiLU activation, and a concluding 3D convolution project the processed features to a single output channel.

\subsection{Network training and inference}
Protein structural embeddings provide an additional modality containing structure information, serving as key-value pairs in the attention mechanism when training. However, since these embeddings are unavailable during validation and inference, we implemented a specialized mode in which the network bypasses the cross-attention operation. In this mode, the network relies exclusively on feedforward transformations with residual connections. This design maintains consistency between training and validation/inference phases while preserving the learned feature representations.

\noindent \textbf{Training} \hspace{3pt}
During training, CryoSAMU accepts an \emph{exp.3D-image} and its corresponding structural embedding as input and generates an enhanced 3D image (denoted as \emph{pred.3D-image}). Previous studies have shown that L1 loss performs well in similar tasks~\cite{deepemhancer, cryofem}. However, to improve training stability in the presence of noisy data and outliers that are common in cryo-EM maps, we employed the smooth L1 loss to encourage the generator to minimize the difference between the output \emph{pred.3D-image}, $X$, and the target 
\emph{tgt.3D-image}, $Y$:
\begin{equation}
    \text{SmoothL1Loss}(X,Y) =
    \begin{cases}
    0.5(X-Y)^2, & \text{if } |X-Y| < 1, \\
    |X - Y| - 0.5, & \text{otherwise}.
    \end{cases}
\end{equation}
Moreover, to enhance the network's robustness, we employed \emph{TorchIO}~\cite{torchio} for data augmentation, including random Gaussian noise, anisotropy, and blurring.

\noindent \textbf{Inference} \hspace{3pt}
During inference, the input experimental map was first zero-padded and divided into smaller cubes ($64\times64\times64$), following the same strategy used for training data. Each cube was then individually processed by the trained neural network to generate the enhanced cube. These enhanced cubes were subsequently reassembled to reconstruct the map as its original dimensions. To prevent loss of spatial information and ensure smooth transition between cubes, only the central $50\times50\times50$ voxels from each enhanced cube were used in the final reconstruction, following the method proposed by Si et al.~\cite{Cascaded-CNN}.

\noindent \textbf{Implementation} \hspace{3pt}
The network was implemented in PyTorch 2.6.0 with CUDA 12.4, running under Python 3.12.8. Training was conducted using a distributed data parallel (DDP) strategy across two computational nodes connected via NVLink, with each node equipped with four NVIDIA A100 GPUs of 40 GB VRAM. This setup supported a maximum batch size of 18 per GPU. The network was trained over 95 epochs, requiring approximately 63 computational hours. The AdamW~\cite{AdamW} optimizer was used with an initial learning rate of 0.0001, along with a cosine annealing learning rate scheduler. To improve training performance while maintaining accuracy, automatic mixed precision training was applied. Additionally, gradient clipping (set to 0.5) was applied to prevent gradient explosion.

\section{Experiments and Results}

\begin{figure*}[t]
  \centering
   \includegraphics[width=\linewidth, trim={0.5cm 19.2cm 4.5cm 0.5cm}, clip]{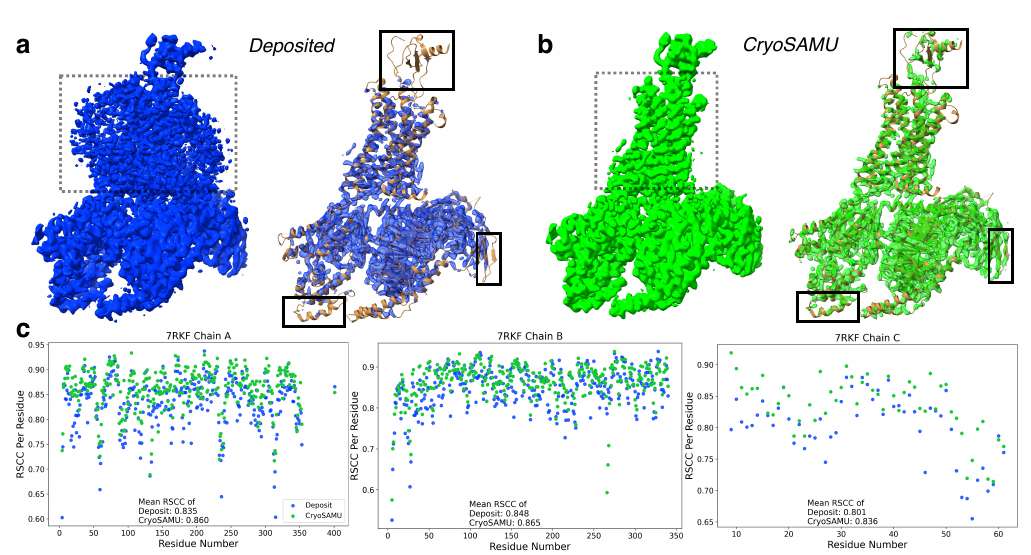}
    \vspace{-15pt}
   \caption{Visual and quantitative comparison of deposited (blue) and CryoSAMU-enhanced (green) maps, with superimposed corresponding PDB structures (brown).
   \textbf{a, b}: Maps are shown at two contour levels. Left: recommended contour level (volume = 85.74e3). Right: higher contour level (volume = 22.57e3).
   \textbf{c}: RSCC comparisons between deposited and CryoSAMU-enhanced maps.
   The example protein is a CX3CL1-US28-G11iN18-scFv16 in TL-state (PDB-7RKF, EMDB-24496, reported resolution of 4.00 {\AA})~\cite{7RKF}. 
   }
   \label{fig:vis_comp}
\end{figure*}

We conducted a comprehensive study to assess the performance of CryoSAMU using a test set of 75 intermediate-resolution cryo-EM density maps and associated PDB structures across a wide range of evaluation metrics.

\subsection{Visualization and quantification of map enhancement}
We first visualized a CryoSAMU-enhanced map alongside its associated deposited map using UCSF ChimeraX \cite{chimerax}. For a fair comparison, both sets of maps were illustrated with the same volume, which requires contour level adjustments owing to differences in their volume ranges. Specifically, we first presented the deposited map at its recommended contour level and volume, then adjusted the contour level of the corresponding CryoSAMU-enhanced map to match the recommended volume. In addition, we also visualized both maps at a higher contour level with the same volume. 

As displayed in Figure \ref{fig:vis_comp}b, CryoSAMU significantly suppressed noise in the lip nanodisc regions (highlighted by dashed boxes in Figure \ref{fig:vis_comp}a) of the deposited map for EMDB-24496 (PDB-7RKF). Moreover, the deposited map at a smaller volumes missed certain structural regions corresponding to the protein structures, as highlighted by black boxes in Figure \ref{fig:vis_comp}a. In contrast, the CryoSAMU-enhanced maps exhibited better alignment with the corresponding protein structures, revealing more structural details, as demonstrated by black boxes in Figure \ref{fig:vis_comp}b.
Similar visual results were observed for another protein structure (see Supplementary Figure \ref{fig:SI_vis_comp}). 
Furthermore, residue-level real-space correlation coefficient (RSCC) measurements~\cite{phenixeval} in Figure \ref{fig:vis_comp}c suggested significant improvements. Specifically, Chains A, B, and C exhibit RSCC increases compared to the deposited map, with correlations rising from 0.835 to 0.860, 0.848 to 0.865, and 0.801 to 0.836, respectively. In addition, 84.9\%, 73.5\%, and 90.6\% of resiudes in Chains A, B, and C, respectively, showcased higher RSCC scores. Consistent RSCC improvements were also observed in other samples (see Supplementary Figure \ref{fig:SI_plot_RSCC}).

\subsection{Benchmark I: improvement of real and Fourier space correlations} \label{bm1}

\begin{table*}[t]
\begin{center}
\begin{tabular}{lcccccc}
\toprule
Metric & Deposit & Autosharpen & DeepEMhancer & EMReady &\textbf{CryoSAMU}(ours) & \begin{tabular}[c]{@{}c@{}}CryoSAMU\\(w/o struct.)\end{tabular} \\
\hline
CC\_box $\uparrow$ & 0.731 & 0.679 & 0.618 & 0.862 & 0.834 & 0.751 \\
CC\_peaks $\uparrow$ & 0.750 & 0.722 & 0.611 & 0.774 & 0.753 & 0.698 \\
CC\_volume $\uparrow$ & 0.594 & 0.542 & 0.534 & 0.729 & 0.691 & 0.571 \\
FSC05 $\downarrow$ & 6.124 & 6.147 & 5.283 & 4.668 & 5.108 & 6.434 \\
\bottomrule
\end{tabular}
\end{center}
\caption{Comparison of different methods across various metrics. See Section \ref{bm1}.}
\label{tab:benchmark}
\end{table*}

\begin{figure*}[ht]
  \centering
   \includegraphics[width=\linewidth, trim={0cm 0cm 0cm 0cm}, clip]{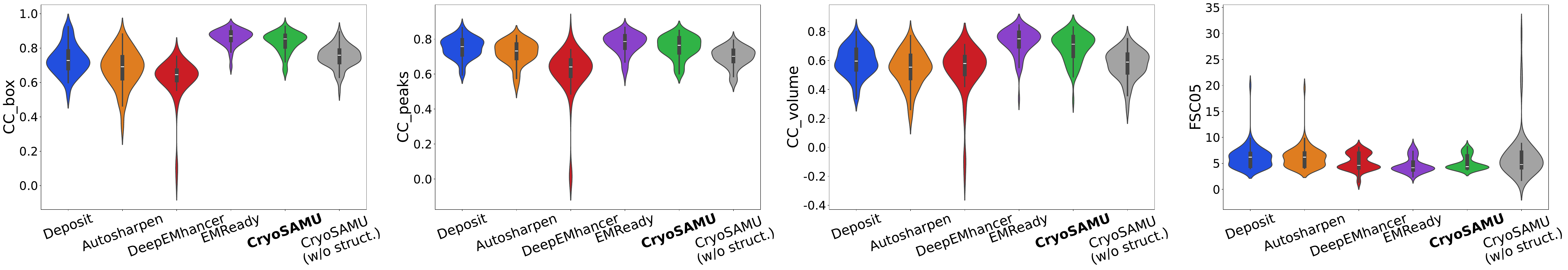}
    \vspace{-15pt}
   \caption{The violin plots for comparison of different methods across four evaluation metrics (see Section \ref{bm1}) over 75 test samples.}
   \label{fig:benchmark}
\end{figure*}

We then benchmarked CryoSAMU against other state-of-the-art methods, including Autosharpen~\cite{phenixautosharpen}, DeepEMhancer~\cite{deepemhancer}, and EMReady~\cite{emready}, in terms of both real-space and reciprocal-space (i.e., Fourier-space) correlations, across a test set of 75 primary maps. 
For real-space correlation, we computed three correlation metrics using \texttt{phenix.map\_model\_cc}~\cite{phenixeval} for each map-model pair (where the model refers to a protein structure): CC\_box, CC\_volume, and CC\_peaks. These metrics differ based on the choice of map regions used in the calculations.
CC\_box considers the entire map. CC\_volume and CC\_peaks focus on regions with the highest density values. However, CC\_volume selects grid points only around atomic centers, while CC\_peaks selects points located anywhere within the volume. For all three metrics, higher values indicate better map performance.
For Fourier-space correlation, we computed Fourier shell correlation (FSC) using \texttt{phenix.mtriage}~\cite{phenixeval}, and reported the unmasked map-model FSC05 values. FSC values are typically represented as a function of the inverse map resolution, where lower value indicates better map resolution.

The average real-space CC and FSC values are listed in Table \ref{tab:benchmark}. 
According to the violin plots shown in Figure \ref{fig:benchmark}, CryoSAMU-enhanced maps demonstrated significant improvements over the deposited maps in terms of CC\_box and CC\_volume, with average values increasing from 0.731 to 0.834 and from 0.594 to 0.691, respectively. The average CC\_peaks score showed a slightly increase from 0.750 to 0.753. These results indicate that CryoSAMU effectively enhances deposited maps in both the entire region and the highest-density regions. In contrast, maps processed by Autosharpen and DeepEMhancer exhibited lower scores across all three metrics. EMReady showed slightly better improvements than CryoSAMU across all three metrics.
For FSC05 scores, CryoSAMU outperformed the deposited map, Autosharpen, and DeepEMhancer, achieving an average value of 5.108 {\AA}. However, it slightly underperformed compared to EMReady, which achieved an average value of 4.668 {\AA}. 
These results demonstrate that both CryoSAMU and EMReady consistently enhance the deposited maps in terms of correlations in both real and Fourier spaces.

\subsection{Benchmark II: improvement of protein structure modeling} \label{sec:bm2}

As the goal of enhancing cryo-EM density maps is to improve the performance of protein structure modeling from density maps (i.e., map interpretability), we benchmarked protein structures constructed  from CryoSAMU-enhanced maps against those processed by other methods.
Specifically, we used a standard structure modeling tool, known as \texttt{phenix.map\_to\_model}~\cite{phenix}, to construct protein structures from 20 maps enhanced by the different tested methods. These maps were randomly selected from the test dataset to ensure that they were not exposed during training, as listed in Supplementary Table \ref{tab:test_data_m2m}. 
To evaluate these structures, we used \texttt{phenix.chain\_comparison}~\cite{phenix} to compare the constructed structures against their corresponding ground-truth PDB protein structures. We reported two metrics: residue coverage and sequence match.
The residue coverage indicates the fraction of residues in the query structure that match the corresponding residues in the target structure within 3.0 {\AA}, regardless of residue type. The sequence match indicates the percentage of matched residues that share identical residue types between the query and target structures.

\begin{table}[ht]
\begin{center}
\begin{tabular}{lcc}
\toprule
Method & \begin{tabular}[c]{@{}c@{}}Residue\\Coverage (\%) $\uparrow$ \end{tabular} & \begin{tabular}[c]{@{}c@{}}Sequence\\Match (\%) $\uparrow$ \end{tabular} \\
\hline
Deposit & 31.71 & 8.42 \\
Autosharpen & 16.00 & 8.13 \\
DeepEMhancer & 24.31 & 10.0 \\
EMReady & 31.61 & 11.38 \\
\textbf{CryoSAMU}(ours) & 38.03 & 9.33 \\
CryoSAMU(w/o struct.) & 8.08 & 8.13 \\
\bottomrule
\end{tabular}
\end{center}
\caption{Comparison of average residue coverage and sequence match across different methods.}
\label{tab:m2m}
\end{table}

\begin{figure*}[ht]
  \centering
   \includegraphics[width=\linewidth, trim={1cm 24cm 0.8cm 0cm}, clip]{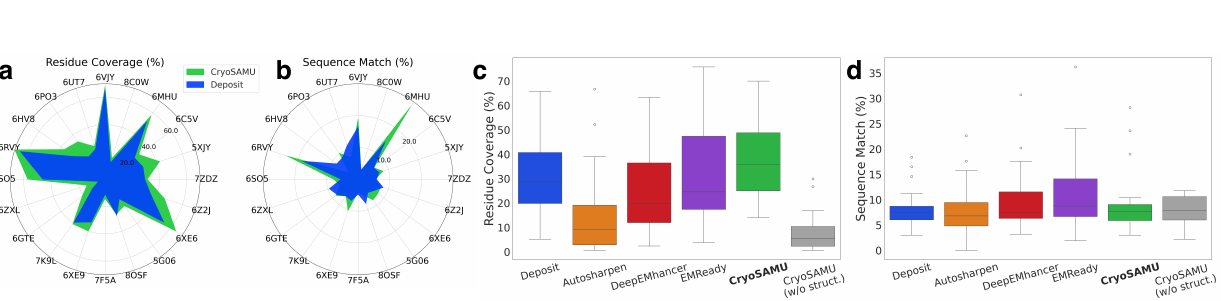}
   \vspace{-15pt}
   \caption{
   \textbf{a-b}: The polar plots for comparison of protein structures constructed from deposited (blue) and CryoSAMU-enhanced (green) maps, using metrics of (\textbf{a}) residue coverage and (\textbf{b}) sequence match.
   \textbf{c-d}: The box-whisker plots for comparison of different methods across two evaluation metrics over 20 test samples. See Section \ref{sec:bm2}.
   }
   \label{fig:fig_m2m}
\end{figure*}

The average metric scores from all methods are listed in Table \ref{tab:m2m}.
Figure \ref{fig:fig_m2m}a and b provide a detailed comparison for each individual test examples in terms of residue coverage and sequence match, respectively. The polar plots clearly showcase that after CryoSAMU enhancement, 19 out of 20 samples exhibited an improvement in residue coverage on deposited maps, with the average score increasing from 31.71\% to 38.03\%; and 55\% of samples exhibited an improvement in sequence match, with the average score increasing from  8.42\% to 9.33\%. 
Furthermore, we benchmarked CryoSAMU against other methods, as shown in Figure \ref{fig:fig_m2m}c and d. CryoSAMU achieved the highest residue coverage score among all methods, although its correlation scores were slightly lower than those of EMReady. The sequence match score of CryoSAMU was slightly lower than EMReady and DeepEMHancer, while still better than the deposited maps.
These results demonstrate that CryoSAMU enhancement boosts protein structure modeling performance.

\subsection{Benchmark III: processing time} \label{bm3}

\begin{figure}[ht]
  \centering
   \includegraphics[width=0.8\linewidth, trim={0cm 0cm 0cm 0cm}, clip]{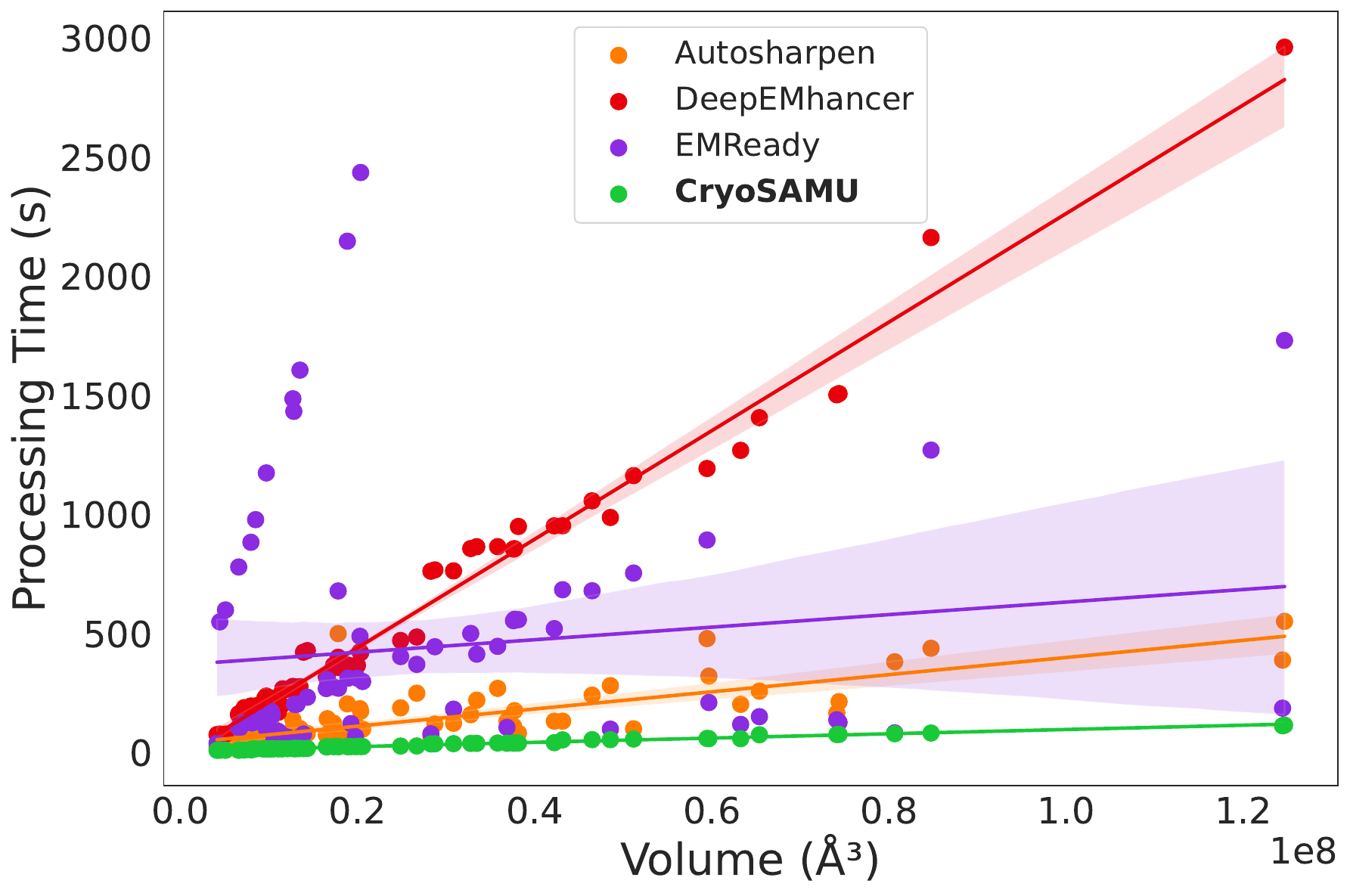}
   \vspace{-15pt}
   \caption{The scatter plot of map processing time against map volume. Each dot represents the processing time for an individual map based on its volume. The shaded area around the regression line denotes the confidence interval of the regression estimate.}
   \label{fig:time}
\end{figure}

To evaluate the scalability of CryoSAMU in practice, we recorded the time required to generate each enhanced map of all 75 test samples and compared it against the processing time of other methods. Figure \ref{fig:time} shows the wall-clock time plotted against the volume size of input experimental maps, ranging from the order of $10^6$ to $10^8$ $\text{Å}^3$.
For a fair comparison, all methods were run on the same workstation equipped with an AMD Ryzen Threadripper 2950X Processor of 32 CPUs and an NVIDIA GeForce RTX 2080 Ti of 12 GB VRAM. Each method was executed with the maximum batch size that our GPU can accommodate: approximately 12 for DeepEMhancer, 64 for EMReady, and 24 for CryoSAMU.
CryoSAMU (shown in green) displayed the minimum processing time across maps of varying volumes. Its weak linear dependency on map volume and tight confidence interval around its fit line indicate that CryoSAMU has both optimal scalability and consistent performance. 
In contrast, DeepEMhancer exhibited a strong linear correlation between processing time and map volume, indicating poor scalability as volume size increases.
EMReady showed a wider confidence interval in its linear fit, reflecting high variability in processing time. Notably, several outliers at lower volumes showed significantly longer processing time compared to other methods.
For a significantly large map with a volume size of $1.25 \times 10^8$ $\text{Å}^3$, CryoSAMU took only 116.48 seconds for generating an enhanced map, while Autosharpen, DeepEMHancer, and EMReady took 552.19, 2963.10, and 1731.719 seconds, respectively.
Table \ref{tab:time} lists the average processing time for each method. CryoSAMU achieved an average processing time of 32.49 seconds, approximately 13.6 times faster than EMReady, while generating comparably enhanced maps. These results suggest that CryoSAMU scales efficiently with increasing map volume, making it a promising tool for practical applications.

\begin{table}[t]
\begin{center}
\setlength{\tabcolsep}{4pt} 
\begin{tabular}{cccc}
\toprule
Autosharpen & DeepEMhancer & EMReady &\textbf{CryoSAMU} \\
\hline
138$\pm$118 & 544$\pm$517 & 441$\pm$500 & 32$\pm$23 \\
\bottomrule
\end{tabular}
\end{center}
\caption{Average processing time in seconds of different methods.}
\label{tab:time}
\end{table}

\subsection{Ablation study}

We finally conducted an ablation study to evaluate the impact of integrating structural modality. We compared CryoSAMU with (w/) and without (w/o) structural embeddings using 75 test samples for correlation evaluation and 20 test samples for protein structure modeling assessment.

\begin{figure}[ht]
  \centering
   \includegraphics[width=\linewidth, trim={0cm 0cm 0cm 0cm}, clip]{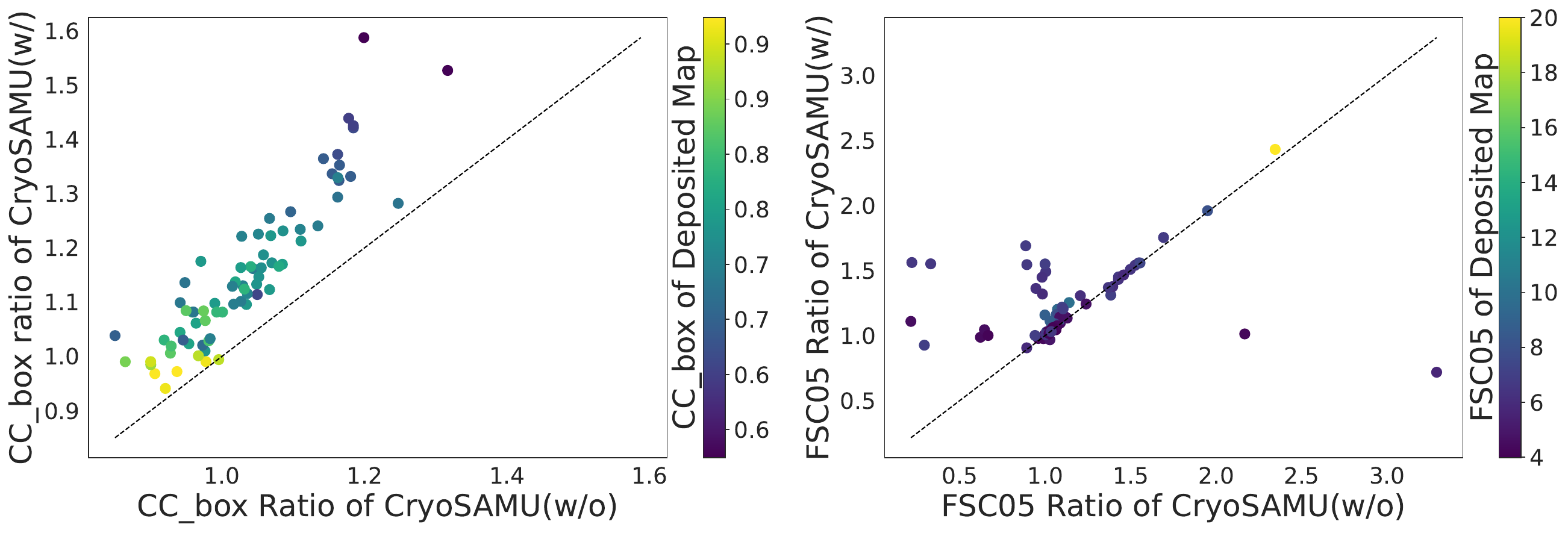}
   \vspace{-15pt}
   \caption{Pairwise comparison of enhanced/deposited ratios for CryoSAMU (w/) and (w/o). Each point represents a single map. Point colors encode the deposited map quality.
   }
   \label{fig:ablation}
\end{figure}

Figure \ref{fig:ablation} shows that CryoSAMU (w/) outperforms CryoSAMU (w/o) in both CC\_box and FSC05, with improvements of 98.7\% and 77.3\%, respectively, as indicated by scatter points above the diagonal line.
Moreover, the most significant gains (points far from 1.0) were observed in poorer‐quality deposited maps (colored in purple), which tend to have lower deposited CC\_box or higher FSC05 values.
Table \ref{tab:benchmark} lists the average real- and Fourier-space metrics, indicating that incorporating structural embeddings derived from ESM-IF1 led to a significant improvement of map enhancement, as also reported in Figure \ref{fig:benchmark}. In terms of protein structure modeling, residue coverage significantly raised from 8.08\% to 38.03\%, while sequence match raised from 8.13\% to 9.33\% with the integration of structural embeddings. This suggests that structural information helps complement map regions with poor resolutions, artifacts, or noise, thereby increasing the completeness (higher residue coverage) and improving accuracy (higher sequence match) during structure modeling. These findings underscore the importance of integrating structural modality to enable the network to develop structural awareness beyond learning solely from 3D density maps.

\section{Conclusion and Discussion}

In this work, we introduce CryoSAMU, the first structure-aware multimodal network for enhancing cryo-EM density maps at intermediate resolution of protein structures. Our approach combines 3D map features with corresponding structural features through cross-attention mechanisms. 
Notably, structural embeddings are used solely as auxiliary supervision during training to guide the model toward learning biologically-grounded and structure-aware representations from the density maps. 
This strategy aligns with the well-established principle of conditional training, where auxiliary modalities, despite not available at inference, are used during training to improve the quality and generalization of learned representations from the primary input \cite{auxiliary}. 
In addition, we develop a self-attention weighting algorithm to produce fixed-size representations of structural embeddings derived from the pretrained ESM-IF1 model, preserving inter-chain and residue relationships while maintaining structural integrity. 
Our benchmark results demonstrate that CryoSAMU preforms competitively with existing cutting-edge methods, closely approaching the performance of EMReady, the current leading tool for cryo-EM density map enhancement. Notably, CryoSAMU achieves the fastest processing speed among all tested methods, positioning it as a promising solution for large-scale and practical applications. Furthermore, our ablation study reveals that incorporating an additional structural modality significantly boosts CryoSAMU's performance across all evaluation metrics, suggesting a new avenue for future cryo-EM research to explore the effective integration of multimodal data during network training.

Despite CryoSAMU demonstrating superior performance in enhancing cryo-EM maps, its current architecture---based on residual convolutions within a U-Net framework---is primarily designed to capture local information. In practice, capturing global context and long-range dependencies across map voxels could further improve performance. This could be addressed by adopting more hierarchical architectures, such as the Swin Transformer~\cite{swintransformer}, which facilitates feature extraction over larger receptive fields. Moreover, incorporating supplementary loss terms, such as the Structural Similarity Index Measure (SSIM) loss, could mitigate overfitting and enhance training efficiency~\cite{emready}. These will be explored in our future work. Furthermore, we plan to expand our dataset by including high-resolution maps, which could increase the robustness of the model and further elevate its performance.

\section*{Acknowledgments}
This work is supported in parts by a NFRFE-2019-00486 grant.

\clearpage
\bibliographystyle{unsrt}  
\bibliography{references}  

\begin{thebibliography}{10}

\bibitem{dd1}
Jean-Paul Renaud, Ashwin Chari, Claudio Ciferri, Wen-ti Liu, Herv{\'e}-William R{\'e}migy, Holger Stark, and Christian Wiesmann.
\newblock Cryo-em in drug discovery: achievements, limitations and prospects.
\newblock {\em Nature reviews Drug discovery}, 17(7):471--492, 2018.

\bibitem{dd2}
Alan Merk, Alberto Bartesaghi, Soojay Banerjee, Veronica Falconieri, Prashant Rao, Mindy~I Davis, Rajan Pragani, Matthew~B Boxer, Lesley~A Earl, Jacqueline~LS Milne, et~al.
\newblock Breaking cryo-em resolution barriers to facilitate drug discovery.
\newblock {\em Cell}, 165(7):1698--1707, 2016.

\bibitem{eminherent}
Peter~B Rosenthal.
\newblock Interpreting the cryo-em map.
\newblock {\em IUCrJ}, 6(1):3--4, 2019.

\bibitem{phenixautosharpen}
Thomas~C Terwilliger, Oleg~V Sobolev, Pavel~V Afonine, and Paul~D Adams.
\newblock Automated map sharpening by maximization of detail and connectivity.
\newblock {\em Biological Crystallography}, 74(6):545--559, 2018.

\bibitem{relionpostprocess}
Dari Kimanius, Bj{\"o}rn~O Forsberg, Sjors~HW Scheres, and Erik Lindahl.
\newblock Accelerated cryo-em structure determination with parallelisation using gpus in relion-2.
\newblock {\em elife}, 5:e18722, 2016.

\bibitem{LocalDeblur}
Erney Ram{\'\i}rez-Aportela, Jose~Luis Vilas, Alisa Glukhova, Roberto Melero, Pablo Conesa, Marta Mart{\'\i}nez, David Maluenda, Javier Mota, Amaya Jim{\'e}nez, Javier Vargas, et~al.
\newblock Automatic local resolution-based sharpening of cryo-em maps.
\newblock {\em Bioinformatics}, 36(3):765--772, 2020.

\bibitem{deepemhancer}
Ruben Sanchez-Garcia, Josue Gomez-Blanco, Ana Cuervo, Jose~Maria Carazo, Carlos Oscar~S Sorzano, and Javier Vargas.
\newblock Deepemhancer: a deep learning solution for cryo-em volume post-processing.
\newblock {\em Communications biology}, 4(1):874, 2021.

\bibitem{cryofem}
Xin Dai, Longlong Wu, Shinjae Yoo, and Qun Liu.
\newblock Integrating alphafold and deep learning for atomistic interpretation of cryo-em maps.
\newblock {\em Briefings in Bioinformatics}, 24(6):bbad405, 2023.

\bibitem{emgan}
Sai~Raghavendra Maddhuri Venkata~Subramaniya, Genki Terashi, and Daisuke Kihara.
\newblock Enhancing cryo-em maps with 3d deep generative networks for assisting protein structure modeling.
\newblock {\em Bioinformatics}, 39(8):btad494, 2023.

\bibitem{emready}
Jiahua He, Tao Li, and Sheng-You Huang.
\newblock Improvement of cryo-em maps by simultaneous local and non-local deep learning.
\newblock {\em Nature Communications}, 14(1):3217, 2023.

\bibitem{LocSpiral}
Satinder Kaur, Josue Gomez-Blanco, Ahmad~AZ Khalifa, Swathi Adinarayanan, Ruben Sanchez-Garcia, Daniel Wrapp, Jason~S McLellan, Khanh~Huy Bui, and Javier Vargas.
\newblock Local computational methods to improve the interpretability and analysis of cryo-em maps.
\newblock {\em Nature communications}, 12(1):1240, 2021.

\bibitem{cnn}
Yann LeCun and Yoshua Bengio.
\newblock Convolutional networks for images, speech, and time series.
\newblock {\em The handbook of brain theory and neural networks}, 3361(10):1995, 1995.

\bibitem{gan}
Antonia Creswell, Tom White, Vincent Dumoulin, Kai Arulkumaran, Biswa Sengupta, and Anil~A Bharath.
\newblock Generative adversarial networks: An overview.
\newblock {\em IEEE signal processing magazine}, 35(1):53--65, 2018.

\bibitem{transformer}
Ashish Vaswani, Noam Shazeer, Niki Parmar, Jakob Uszkoreit, Llion Jones, Aidan~N Gomez, {\L}ukasz Kaiser, and Illia Polosukhin.
\newblock Attention is all you need.
\newblock {\em Advances in neural information processing systems}, 30, 2017.

\bibitem{esmif}
Chloe Hsu, Robert Verkuil, Jason Liu, Zeming Lin, Brian Hie, Tom Sercu, Adam Lerer, and Alexander Rives.
\newblock Learning inverse folding from millions of predicted structures.
\newblock In {\em International conference on machine learning}, pages 8946--8970. PMLR, 2022.

\bibitem{swintransformer}
Ze~Liu, Yutong Lin, Yue Cao, Han Hu, Yixuan Wei, Zheng Zhang, Stephen Lin, and Baining Guo.
\newblock Swin transformer: Hierarchical vision transformer using shifted windows.
\newblock In {\em Proceedings of the IEEE/CVF international conference on computer vision}, pages 10012--10022, 2021.

\bibitem{esm1}
Alexander Rives, Joshua Meier, Tom Sercu, Siddharth Goyal, Zeming Lin, Jason Liu, Demi Guo, Myle Ott, C~Lawrence Zitnick, Jerry Ma, et~al.
\newblock Biological structure and function emerge from scaling unsupervised learning to 250 million protein sequences.
\newblock {\em Proceedings of the National Academy of Sciences}, 118(15):e2016239118, 2021.

\bibitem{esm2}
Zeming Lin, Halil Akin, Roshan Rao, Brian Hie, Zhongkai Zhu, Wenting Lu, Nikita Smetanin, Robert Verkuil, Ori Kabeli, Yaniv Shmueli, et~al.
\newblock Evolutionary-scale prediction of atomic-level protein structure with a language model.
\newblock {\em Science}, 379(6637):1123--1130, 2023.

\bibitem{esm3}
Thomas Hayes, Roshan Rao, Halil Akin, Nicholas~J Sofroniew, Deniz Oktay, Zeming Lin, Robert Verkuil, Vincent~Q Tran, Jonathan Deaton, Marius Wiggert, et~al.
\newblock Simulating 500 million years of evolution with a language model.
\newblock {\em Science}, page eads0018, 2025.

\bibitem{proteinbert}
Nadav Brandes, Dan Ofer, Yam Peleg, Nadav Rappoport, and Michal Linial.
\newblock Proteinbert: a universal deep-learning model of protein sequence and function.
\newblock {\em Bioinformatics}, 38(8):2102--2110, 2022.

\bibitem{prostt5}
Michael Heinzinger, Konstantin Weissenow, Joaquin~Gomez Sanchez, Adrian Henkel, Milot Mirdita, Martin Steinegger, and Burkhard Rost.
\newblock Bilingual language model for protein sequence and structure.
\newblock {\em bioRxiv}, pages 2023--07, 2023.

\bibitem{proteinmpnn}
Justas Dauparas, Ivan Anishchenko, Nathaniel Bennett, Hua Bai, Robert~J Ragotte, Lukas~F Milles, Basile~IM Wicky, Alexis Courbet, Rob~J de~Haas, Neville Bethel, et~al.
\newblock Robust deep learning--based protein sequence design using proteinmpnn.
\newblock {\em Science}, 378(6615):49--56, 2022.

\bibitem{alphafold2}
John Jumper, Richard Evans, Alexander Pritzel, Tim Green, Michael Figurnov, Olaf Ronneberger, Kathryn Tunyasuvunakool, Russ Bates, Augustin {\v{Z}}{\'\i}dek, Anna Potapenko, et~al.
\newblock Highly accurate protein structure prediction with alphafold.
\newblock {\em nature}, 596(7873):583--589, 2021.

\bibitem{EMDB}
Catherine~L Lawson, Ardan Patwardhan, Matthew~L Baker, Corey Hryc, Eduardo~Sanz Garcia, Brian~P Hudson, Ingvar Lagerstedt, Steven~J Ludtke, Grigore Pintilie, Raul Sala, et~al.
\newblock Emdatabank unified data resource for 3dem.
\newblock {\em Nucleic acids research}, 44(D1):D396--D403, 2016.

\bibitem{PDB}
Helen~M Berman, Tammy Battistuz, Talapady~N Bhat, Wolfgang~F Bluhm, Philip~E Bourne, Kyle Burkhardt, Zukang Feng, Gary~L Gilliland, Lisa Iype, Shri Jain, et~al.
\newblock The protein data bank.
\newblock {\em Acta Crystallographica Section D: Biological Crystallography}, 58(6):899--907, 2002.

\bibitem{chimerax}
Eric~F Pettersen, Thomas~D Goddard, Conrad~C Huang, Gregory~S Couch, Daniel~M Greenblatt, Elaine~C Meng, and Thomas~E Ferrin.
\newblock Ucsf chimera—a visualization system for exploratory research and analysis.
\newblock {\em Journal of computational chemistry}, 25(13):1605--1612, 2004.

\bibitem{tempy2}
Tristan Cragnolini, Harpal Sahota, Agnel~Praveen Joseph, Aaron Sweeney, Sony Malhotra, Daven Vasishtan, and Maya Topf.
\newblock Tempy2: a python library with improved 3d electron microscopy density-fitting and validation workflows.
\newblock {\em Acta Crystallographica Section D: Structural Biology}, 77(1):41--47, 2021.

\bibitem{embuild}
Jiahua He, Peicong Lin, Ji~Chen, Hong Cao, and Sheng-You Huang.
\newblock Model building of protein complexes from intermediate-resolution cryo-em maps with deep learning-guided automatic assembly.
\newblock {\em Nature Communications}, 13(1):4066, 2022.

\bibitem{silu}
Stefan Elfwing, Eiji Uchibe, and Kenji Doya.
\newblock Sigmoid-weighted linear units for neural network function approximation in reinforcement learning.
\newblock {\em Neural networks}, 107:3--11, 2018.

\bibitem{linearattention}
Angelos Katharopoulos, Apoorv Vyas, Nikolaos Pappas, and Fran{\c{c}}ois Fleuret.
\newblock Transformers are rnns: Fast autoregressive transformers with linear attention.
\newblock In {\em International conference on machine learning}, pages 5156--5165. PMLR, 2020.

\bibitem{torchio}
Fernando P{\'e}rez-Garc{\'\i}a, Rachel Sparks, and S{\'e}bastien Ourselin.
\newblock Torchio: a python library for efficient loading, preprocessing, augmentation and patch-based sampling of medical images in deep learning.
\newblock {\em Computer Methods and Programs in Biomedicine}, 208:106236, 2021.

\bibitem{Cascaded-CNN}
Dong Si, Spencer~A Moritz, Jonas Pfab, Jie Hou, Renzhi Cao, Liguo Wang, Tianqi Wu, and Jianlin Cheng.
\newblock Deep learning to predict protein backbone structure from high-resolution cryo-em density maps.
\newblock {\em Scientific reports}, 10(1):4282, 2020.

\bibitem{AdamW}
Ilya Loshchilov and Frank Hutter.
\newblock Decoupled weight decay regularization.
\newblock {\em arXiv preprint arXiv:1711.05101}, 2017.

\bibitem{7RKF}
Naotaka Tsutsumi, Shoji Maeda, Qianhui Qu, Martin V{\"o}gele, Kevin~M Jude, Carl-Mikael Suomivuori, Ouliana Panova, Deepa Waghray, Hideaki~E Kato, Andrew Velasco, et~al.
\newblock Atypical structural snapshots of human cytomegalovirus gpcr interactions with host g proteins.
\newblock {\em Science advances}, 8(3):eabl5442, 2022.

\bibitem{phenixeval}
Pavel~V Afonine, Bruno~P Klaholz, Nigel~W Moriarty, Billy~K Poon, Oleg~V Sobolev, Thomas~C Terwilliger, Paul~D Adams, and Alexandre Urzhumtsev.
\newblock New tools for the analysis and validation of cryo-em maps and atomic models.
\newblock {\em Acta Crystallographica Section D: Structural Biology}, 74(9):814--840, 2018.

\bibitem{phenix}
Thomas~C Terwilliger, Paul~D Adams, Pavel~V Afonine, and Oleg~V Sobolev.
\newblock A fully automatic method yielding initial models from high-resolution cryo-electron microscopy maps.
\newblock {\em Nature methods}, 15(11):905--908, 2018.

\bibitem{auxiliary}
Yu~Shen, Xijun Wang, Peng Gao, and Ming Lin.
\newblock Auxiliary modality learning with generalized curriculum distillation.
\newblock In {\em International Conference on Machine Learning}, pages 31057--31076. PMLR, 2023.

\bibitem{6XE6}
Hongwen Chen, Yang Liu, and Xiaochun Li.
\newblock Structure of human dispatched-1 provides insights into hedgehog ligand biogenesis.
\newblock {\em Life science alliance}, 3(8), 2020.

\bibitem{6M6Z}
Chunfu Xu, Peilong Lu, Tamer~M Gamal El-Din, Xue~Y Pei, Matthew~C Johnson, Atsuko Uyeda, Matthew~J Bick, Qi~Xu, Daohua Jiang, Hua Bai, et~al.
\newblock Computational design of transmembrane pores.
\newblock {\em Nature}, 585(7823):129--134, 2020.

\bibitem{5G06}
Jun-Jie Liu, Chu-Ya Niu, Yao Wu, Dan Tan, Yang Wang, Ming-Da Ye, Yang Liu, Wenwei Zhao, Ke~Zhou, Quan-Sheng Liu, et~al.
\newblock Cryoem structure of yeast cytoplasmic exosome complex.
\newblock {\em Cell research}, 26(7):822--837, 2016.

\end{thebibliography}

\appendix
\newpage
\section*{Supplementary Material}

\renewcommand{\thetable}{S\arabic{table}}
\renewcommand{\thefigure}{S\arabic{figure}}
\setcounter{table}{0} 
\setcounter{figure}{0}

\begin{figure*}[ht]
  \centering
   \includegraphics[width=\linewidth, trim={0cm 21cm 4cm 0cm}, clip]{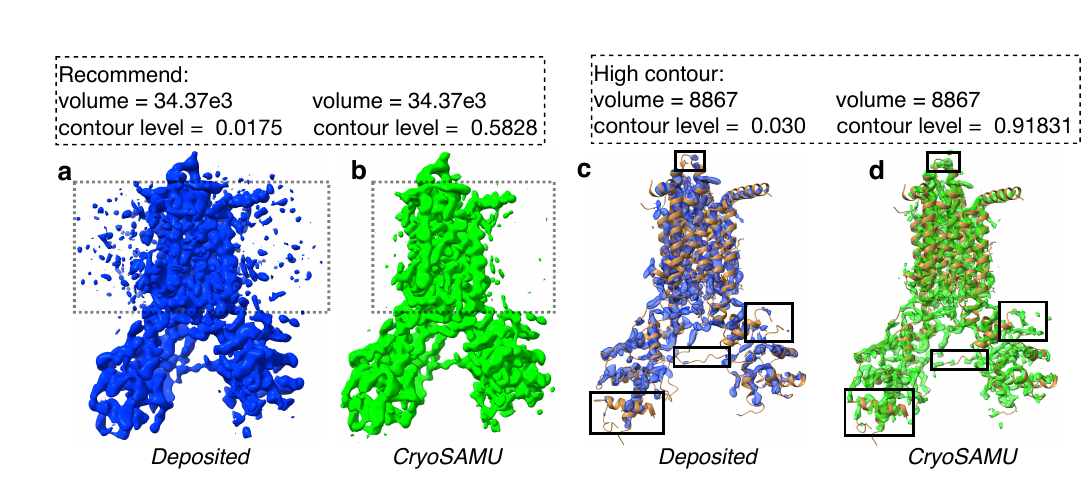}
   \caption{Visualizations of deposited (blue) and CryoSAMU-enhanced (green) maps. The corresponding PDB structures (brown) are superimposed on the maps.
   \textbf{a} Human Dispatched-1 (PDB-6XE6, EMDB-22144, reported resolution of 4.53 {\AA})~\cite{6XE6}. \textbf{a-b}: Maps displayed at the recommended contour level.
   \textbf{c-d}: Maps displayed at a higher contour level. 
   Visualizations were produced by UCSF ChimeraX \cite{chimerax}.
   The protein structure modeling completeness and accuracy improved after CryoSAMU enhancement. For instance, residue coverage increased from 53.5\% to 64.5\%, as well as sequence match increased from 6.1\% to 7.7\%. 
   }
   \label{fig:SI_vis_comp}
\end{figure*}

\clearpage
\newpage

\begin{figure*}[ht]
  \centering
   \includegraphics[width=\linewidth, trim={0cm 0cm 0cm 0cm}, clip]{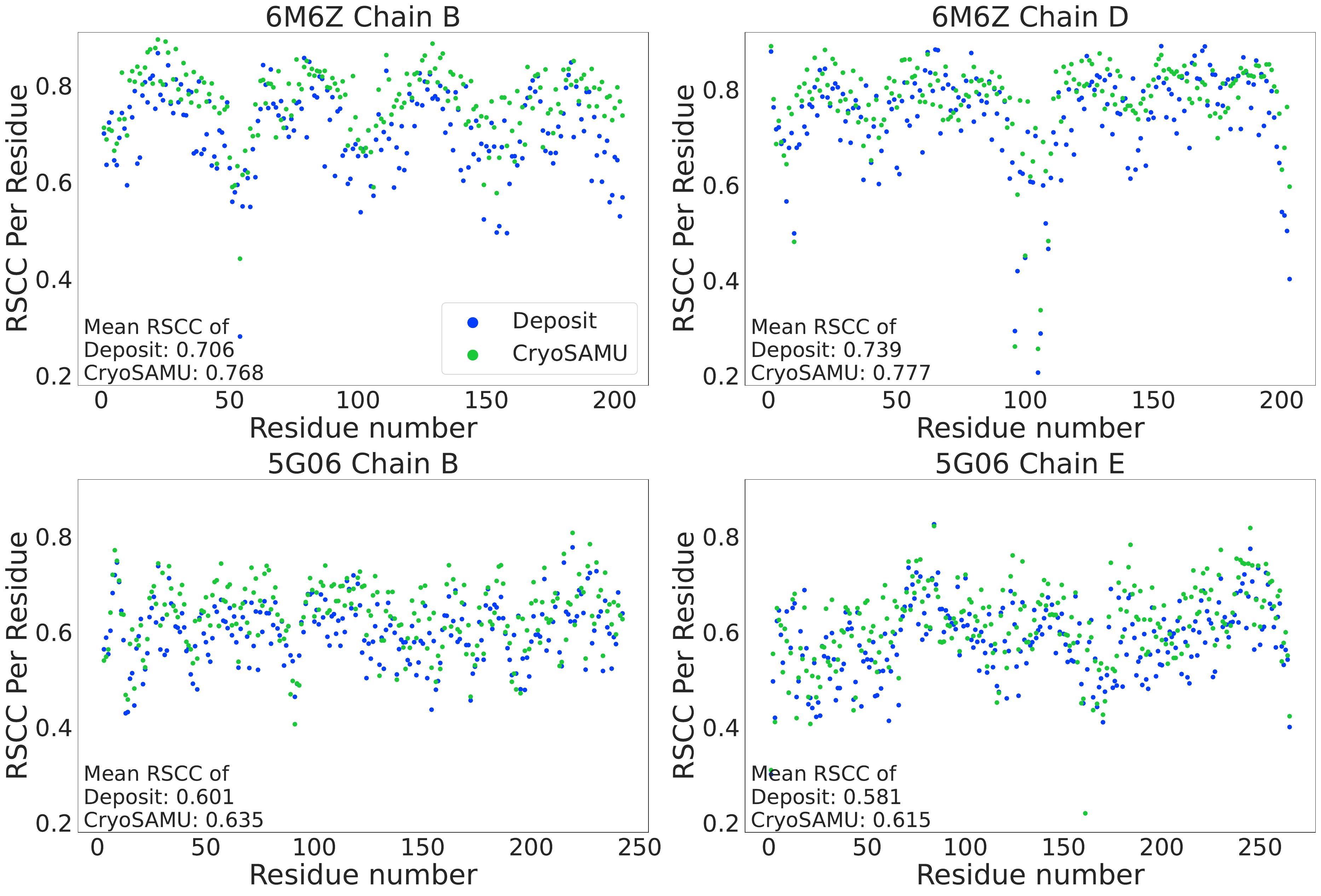}
   \caption{The real-space correlation coefficient (RSCC) comparison between deposited and CryoSAMU-enhanced maps. Top: a transmembrane nanopore TMH4C4 (PDB-6M6Z, EMDB-30126, reported resolution of 5.9 {\AA})~\cite{6M6Z}. Bottom: a yeast cytoplasmic exosome (PDB-5G06, EMDB-3366, reported resolution of 4.2 {\AA})~\cite{5G06}.
   In the first example of PDB-6M6Z, both Chain B and Chain D exhibited significant RSCC improvements compared to the deposited maps, increasing from 0.706 to 0.768 and from 0.739 to 0.777, respectively. In addition, 85.2\% of residues in Chain B and 70.9\% of residues in Chain D showed an increase in their RSCC scores. In the second example of PDB-5G06, the average RSCC scores increased from 0.601 to 0.635 for Chain B, with 82.9\% of 240 residues showing improvement; and increased from 0.581 to 0.615 for Chain E, with 76.2\% of 265 residues perform better.
   }
   \label{fig:SI_plot_RSCC}
\end{figure*}

\clearpage
\newpage

\begin{table*}[p]
  \centering
   \includegraphics[width=0.75\linewidth, trim={0cm 2.5cm 6cm 0cm}, clip]{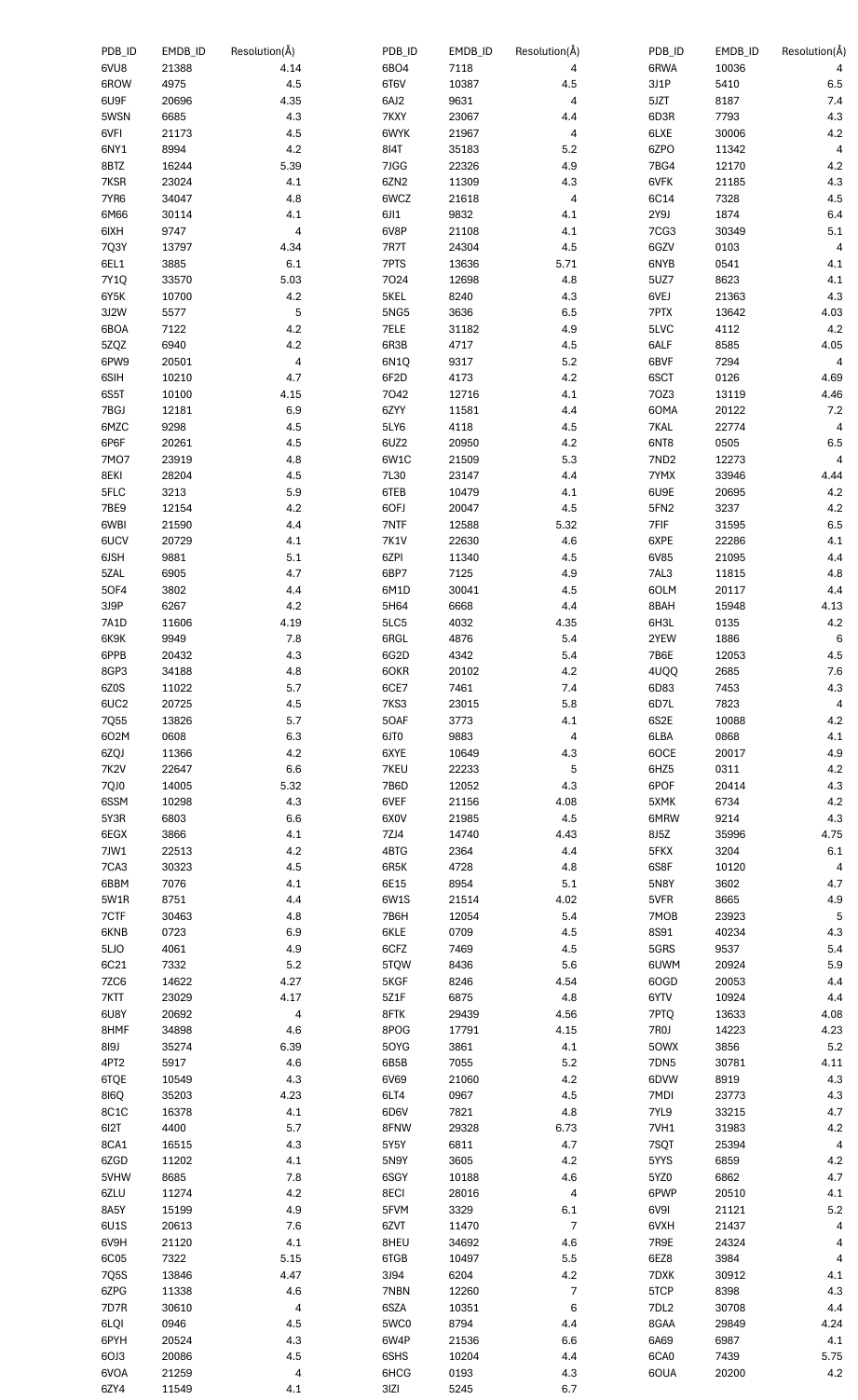}
   \caption{List of all EMDB/PDB examples in training sets.}
   \label{tab:train_data}
\end{table*}

\begin{table*}[p]
  \centering
   \includegraphics[width=0.95\linewidth, trim={0cm 7cm 3cm 0cm}, clip]{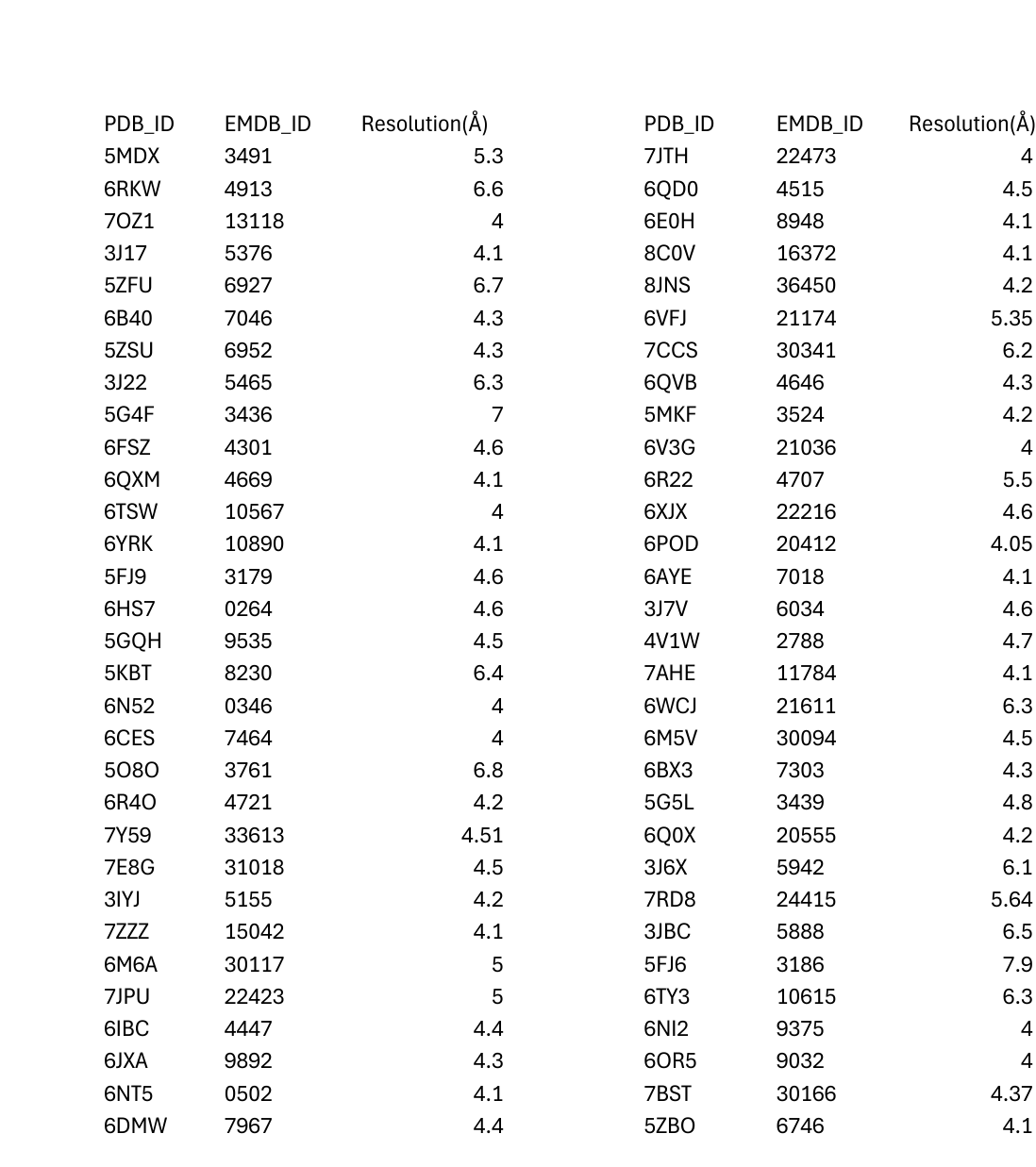}
   \caption{List of all EMDB/PDB examples in validation sets.}
   \label{tab:val_data}
\end{table*}

\begin{table*}[p]
  \centering
   \includegraphics[width=0.9\linewidth, trim={0cm 3cm 3cm 0cm}, clip]{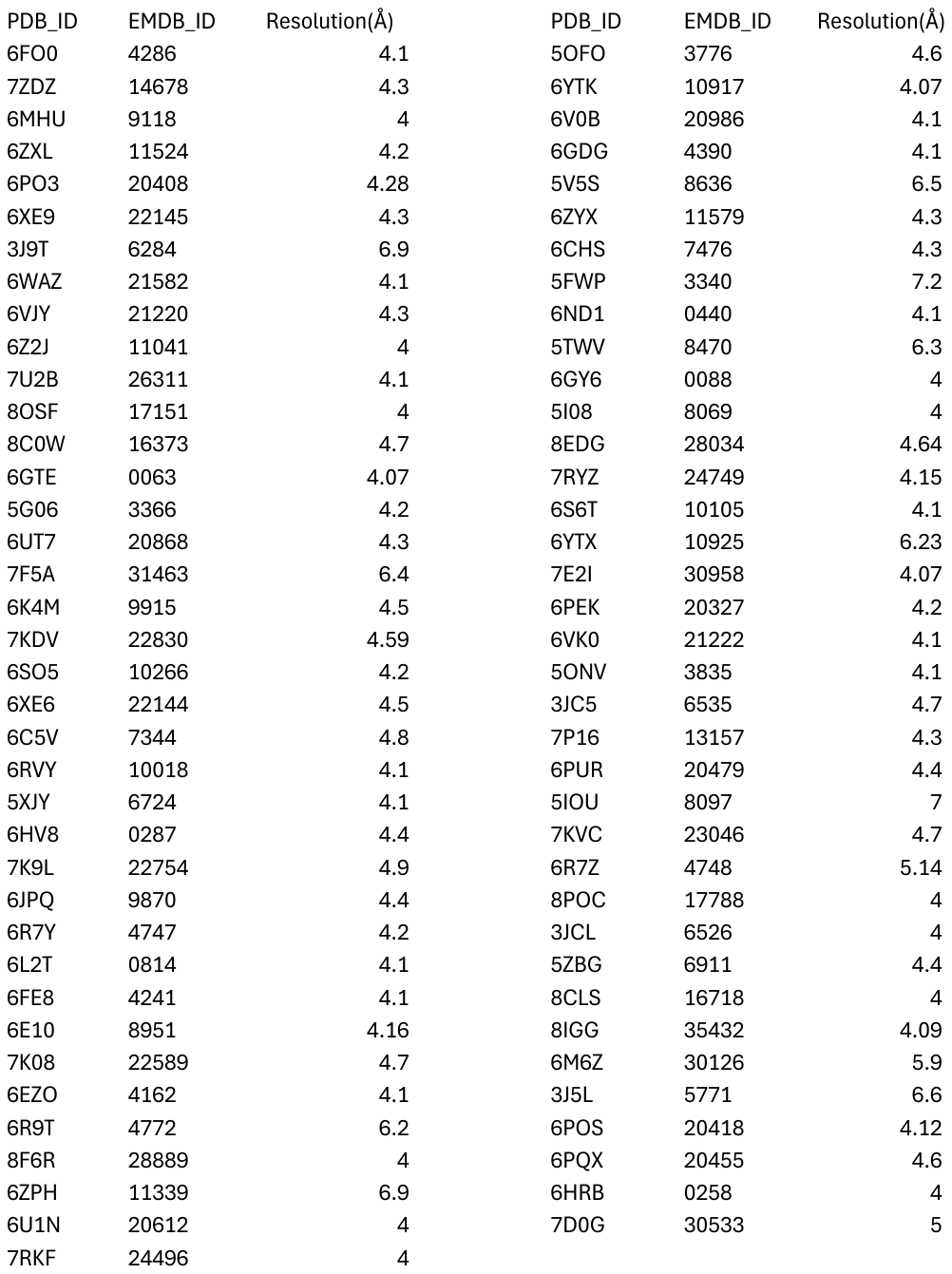}
   \caption{List of all EMDB/PDB examples in test sets.}
   \label{tab:test_data}
\end{table*}

\begin{table*}[p]
  \centering
   \includegraphics[width=0.5\linewidth, trim={0cm 13.5cm 12cm 0cm}, clip]{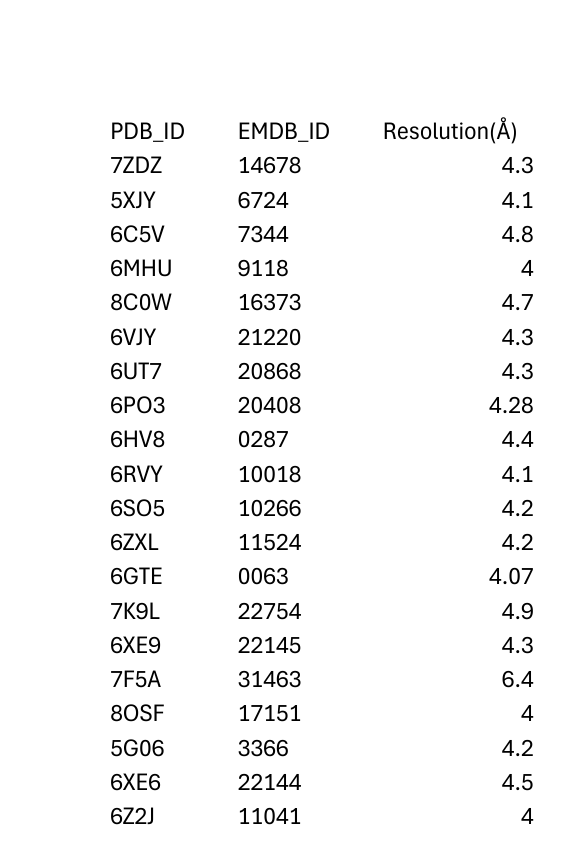}
   \caption{List of all EMDB/PDB examples for protein structure modeling.}
   \label{tab:test_data_m2m}
\end{table*}

\end{document}